\documentclass[11pt]{article}

\usepackage[preprint]{acl}

\usepackage{times}
\usepackage{latexsym}
\usepackage[T1]{fontenc}
\usepackage[utf8]{inputenc}
\usepackage{microtype}
\usepackage{inconsolata}
\usepackage{graphicx}
\usepackage{booktabs}
\usepackage{amsmath}
\usepackage{amssymb}
\usepackage{xcolor}
\usepackage{multirow}

\title{From Pixels to BFS: High Maze Accuracy\\Does Not Imply Visual Planning}

\author{Alberto Rodr\'iguez Salgado Gonzalo \\
  Independent Researcher \\
  \texttt{alberto.rodriguez.salgado.97@gmail.com}}

\usepackage{stfloats}

\makeatletter
\AtBeginDocument{\def\@maketitle{\vbox{\hsize\textwidth
 \linewidth\hsize \vskip 0.125in minus 0.125in \centering
 {\Large\bfseries \@title \par} \vskip 0.18in
 {\def\and{\unskip\enspace{\rmfamily and}\enspace}%
  \def\And{\end{tabular}\hss \egroup \hskip 1in plus 2fil
           \hbox to 0pt\bgroup\hss \begin{tabular}[t]{c}\bfseries}%
  \def\AND{\end{tabular}\hss\egroup \hfil\hfil\egroup
          \vskip 0.25in
           \hbox to \linewidth\bgroup\large \hfil\hfil
             \hbox to 0pt\bgroup\hss \begin{tabular}[t]{c}\bfseries}
  \hbox to \linewidth\bgroup\large \hfil\hfil
    \hbox to 0pt\bgroup\hss
  \outauthor
   \hss\egroup
    \hfil\hfil\egroup}
  \vskip 0.16in
  \begin{center}
    \includegraphics[width=0.96\textwidth]{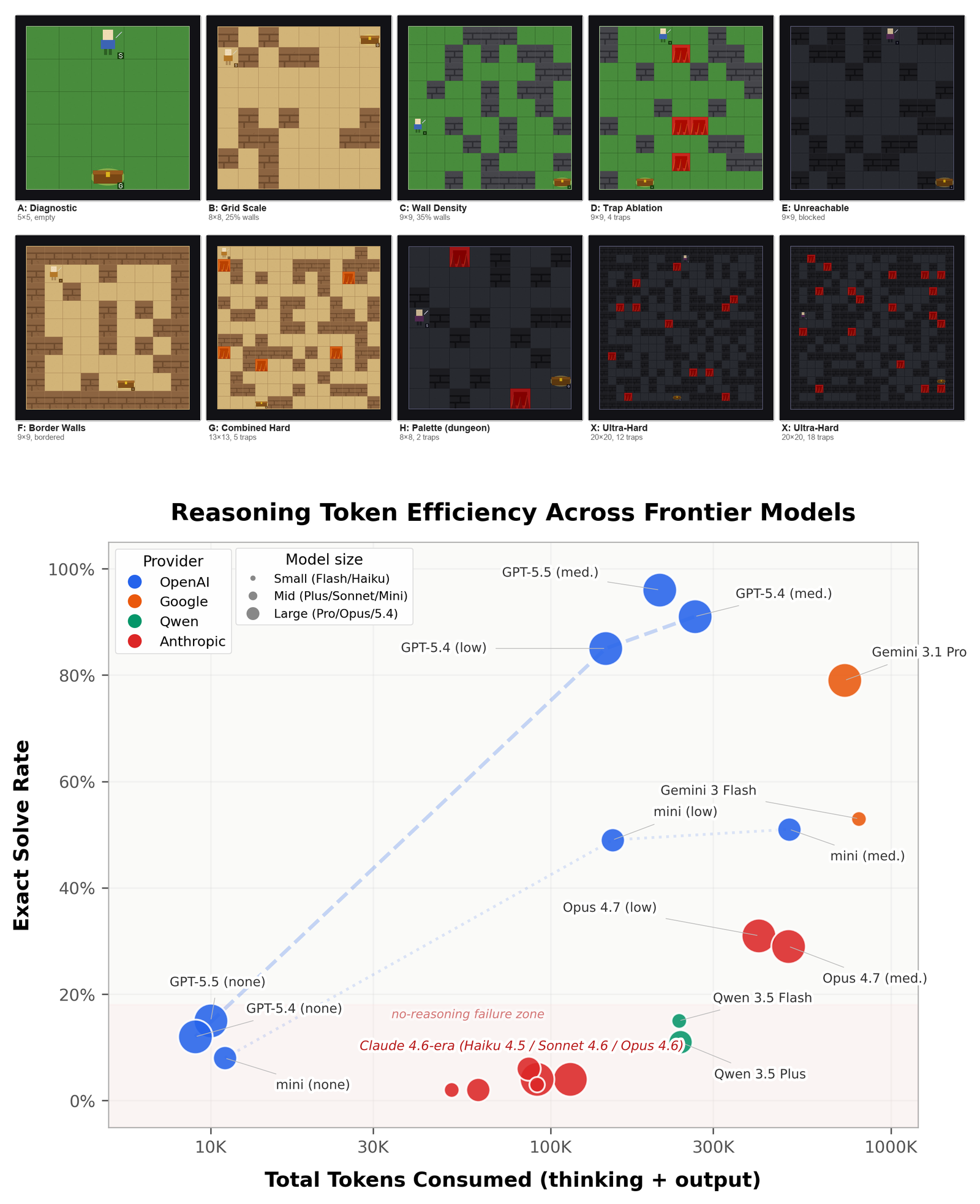}
    \vspace{4pt}
    \parbox{\textwidth}{\fontsize{10pt}{12pt}\selectfont \textbf{Figure 1:} \textbf{Top:} Mazes from \textsc{MazeBench} at increasing difficulty: 5$\times$5 (A) to 20$\times$20 (X). \textbf{Bottom:} Solve rate vs.\ total tokens. Color = provider; size = reasoning effort. The newly released GPT-5.5 sets a new Pareto point (92\% at 1,096 tokens/solve), while the Claude~4.6-era models (red) cluster at 2--6\% and Claude Opus~4.7 jumps to 29--31\%---still trailing the frontier on image input.}
  \end{center}
  \vskip 0.08in
}}}
\makeatother

\begin{document}
\maketitle
\setcounter{figure}{1}

\begin{abstract}
How do multimodal models solve visual spatial tasks---through genuine planning, or through brute-force search in token space?
We introduce \textsc{MazeBench}, a benchmark of 110 procedurally generated maze images across nine controlled groups, and evaluate 19 model configurations from OpenAI, Anthropic, Google, and Alibaba.
The newly released GPT-5.5 solves 96\% of the 100-maze core set and GPT-5.4 91\%, but these scores are misleading: models typically translate images into text grids and then enumerate paths step by step, consuming 1,096--22,818 tokens per solve for a task humans do quickly.
Without added reasoning budgets, all configurations score only 2--15\%; on 20$\times$20 ultra-hard mazes, even GPT-5.5~medium hits token limits and solves only 4/10.
Qualitative traces reveal a common two-stage strategy: image-to-grid translation followed by token-level search, effectively BFS in prose.
A text-grid ablation shows Claude Sonnet~4.6 rising from 6\% to 80\% and the newer Claude Opus~4.7 rising from 31\% to 90\% when given the correct grid, isolating weak visual extraction from downstream search across two Claude generations.
When explicitly instructed not to construct a grid or perform graph search, models still revert to the same enumeration strategy.
\textsc{MazeBench} therefore shows that high accuracy on visual planning tasks does not imply human-like spatial understanding.
\end{abstract}

\section{Introduction}

Multimodal large language models (MLLMs) have achieved impressive performance across a wide range of vision-language tasks, from visual question answering to diagram understanding and mathematical reasoning in visual contexts \cite{lu2024mathvista, yue2024mmmu, fu2024mme}.
Recent benchmarks have begun probing deeper visual capabilities, finding that models struggle with tasks that humans---even young children---solve effortlessly \cite{fu2024blink, tong2024eyes, chen2026babyvision}.
Yet when models \emph{do} score well on such tasks, a crucial question remains: \textbf{does a high accuracy score mean the model actually understands the task, or is it achieving the right answer through a fundamentally different---and far less efficient---mechanism?}

We investigate this question through visual maze solving, a task that is conceptually simple for humans: given a pixel-art maze image, find the shortest path from the player to the treasure.
A human glances at even a complex 20$\times$20 maze and traces the path visually in seconds.
We find that frontier MLLMs can also solve many of these mazes---GPT-5.5 achieves 96\%, GPT-5.4 91\%, Gemini~3.1~Pro 79\%---but they do so in a qualitatively different way.
Rather than spatial planning, models translate the image into a token-level grid representation and then perform \emph{serial path enumeration}: they brute-force the solution step by step in natural language, consuming thousands of reasoning tokens for a task that requires no deliberation for a human.
When the path is too long to enumerate within the token budget, the model gives up---not because it cannot see the maze, but because it runs out of space to think.

This finding has implications beyond mazes.
It suggests that benchmark accuracy alone can be misleading about the nature of model capabilities: a model may score 90\% on a task while using a fundamentally different---and far more costly---cognitive strategy than the one the benchmark was designed to measure.

We design \textsc{MazeBench} with three properties that make this analysis possible:

\paragraph{Controlled difficulty via procedural generation.}
We build a procedural maze generator producing mazes with controlled grid size (5$\times$5 to 20$\times$20), wall density (0--55\%), trap count (0--25), border walls, and varied start/goal positions.
Ground-truth shortest paths are computed via BFS.
The 110 mazes are organized into nine experimental groups---including diagnostic, grid scale, wall density, trap ablation, and ultra-hard---enabling clean ablation studies (Figure~1).

\paragraph{Reasoning effort as a controlled variable.}
We systematically vary the reasoning budget via frontier API controls (OpenAI's \texttt{reasoning\_effort}, Anthropic's adaptive thinking), producing a scaling curve from no thinking to medium effort on the \emph{same} visual inputs.

\paragraph{Token efficiency as a window into strategy.}
We report the total tokens consumed (thinking + output) per solve.
This reveals \emph{how} models solve mazes, not just whether they do: GPT-5.5~low requires 1,096 tokens per solve, GPT-5.4~low 1,710, and Gemini~3~Flash 15,171---all reaching correct answers but through vastly different amounts of brute-force enumeration (Figure~1).

Our main findings are:

\begin{enumerate}
\itemsep0em
\item \textbf{High scores mask brute-force strategies.} GPT-5.5 solves 96\% of mazes and GPT-5.4 91\%, but each consumes thousands of tokens per solve (2,172 and 2,913 respectively) in serial enumeration, with no observable spatial planning behavior. On 20$\times$20 mazes where paths exceed the token budget, even GPT-5.5~medium drops to 4/10 (Section~\ref{sec:ultrahard}).

\item \textbf{Without added reasoning budgets, performance remains very low.} All 19 configurations score only 2--15\% in their no-thinking or lowest-budget settings, even when some parse the grid correctly. For the stronger models, the failure is primarily in path planning rather than basic visual parsing (Section~\ref{sec:diagnostic}).

\item \textbf{The same task, radically different costs.} Models vary by 14$\times$ in tokens per solve (1,096 for GPT-5.5~low vs.\ 15,171 for Gemini~3~Flash), revealing that ``solving a maze'' means very different things computationally across providers (Section~\ref{sec:efficiency}).

\item \textbf{Within a generation, size alone does not buy spatial reasoning.} Within the Claude~4.6 generation, Opus~4.6 (the largest) solves the same 4\% as Haiku~4.5 (the smallest). The newer Claude Opus~4.7 closes much of the gap to frontier models (31\% at low effort), and GPT-5.5 advances the OpenAI ceiling from 91\% to 96\% while cutting tokens-per-solve at low effort by 36\%---showing that generational improvements, not raw scale within a fixed generation, are what actually move this benchmark (Section~\ref{sec:analysis}).
\end{enumerate}

\section{Related Work}

\paragraph{Visual perception gaps in MLLMs.}
Several recent benchmarks have documented systematic failures in visual perception.
BLINK \cite{fu2024blink} reformats 14 classic computer vision tasks as multiple-choice questions and finds that GPT-4V achieves only 51\% versus 96\% for humans, concluding that perception tasks ``resist mediation through natural language.''
\citet{tong2024eyes} identify ``CLIP-blind pairs''---images that vision encoders conflate despite clear visual differences---and construct the MMVP benchmark exposing failures on basic visual patterns.
BabyVision \cite{chen2026babyvision} tests core visual abilities that human children master by age 3--6, finding that even Gemini~3~Pro scores only 49.7 versus 94.1 for adults.
Notably, BabyVision includes a maze-tracing task in which models select which entrance connects to an exit from a multiple-choice list---a \emph{perceptual tracking} task.
Our benchmark differs fundamentally: rather than choosing among predefined options, models must \emph{generate} the complete shortest path as an exact sequence of moves (U/D/L/R), requiring both visual parsing and multi-step spatial planning.
This distinction allows us to show that failures compound across \emph{both} stages: some models (Claude) fail primarily at visual grid extraction, while others (GPT-5.4, Gemini) parse the grid correctly but still resort to brute-force token enumeration rather than spatial planning.

\paragraph{Multimodal reasoning benchmarks.}
MathVista \cite{lu2024mathvista} and MathVerse \cite{zhang2024mathverse} evaluate mathematical reasoning in visual contexts, finding that models often rely on textual cues rather than diagram understanding.
EMMA \cite{hao2025emma} tests cross-modal reasoning across math, physics, chemistry, and coding, reporting that even chain-of-thought prompting and test-time compute scaling underperform (ICML~2025).
MMMU \cite{yue2024mmmu} and MME \cite{fu2024mme} provide comprehensive evaluation suites spanning dozens of disciplines, while MMEvalPro \cite{huang2025mmevalpro} addresses systematic biases in multiple-choice evaluation by introducing perception prerequisite questions.
\citet{chen2024right} audit evaluation methodology itself, questioning whether current benchmarks measure the capabilities they claim to.
Our benchmark complements this body of work by targeting a single, tightly controlled task---visual pathfinding---that isolates spatial reasoning from domain knowledge.

\paragraph{Spatial reasoning in MLLMs.}
Spatial reasoning has emerged as a key evaluation axis for multimodal models.
SpatialVLM \cite{chen2024spatialvlm} endows VLMs with metric spatial reasoning via synthetic data, while SpatialRGPT \cite{cheng2024spatialrgpt} grounds spatial reasoning in depth-aware representations.
SpatialBench \cite{xu2025spatialbench} decomposes spatial intelligence into five cognitive levels and finds that models fail at high-level planning.
SpatiaLab \cite{tong2026spatialab} evaluates spatial reasoning in unconstrained real-world images, finding that even GPT-5-mini scores only 41\% versus 65\% for humans.
GSR-Bench \cite{rajabi2024gsrbench} evaluates grounded spatial relationship understanding across 27~models at NeurIPS~2024.
VGRP-Bench \cite{ren2025vgrpbench} is the most directly related benchmark to ours, testing vision-language models on grid-based visual reasoning puzzles.
Our work differs in three ways: (1)~we introduce reasoning effort as a controlled experimental variable, (2)~we report thinking token efficiency as a metric, and (3)~we demonstrate through qualitative analysis that models solve grid puzzles through brute-force token-level enumeration, with no evidence of human-like spatial planning.

\paragraph{Test-time compute scaling.}
\citet{snell2025scaling} demonstrate that scaling inference-time computation can be more effective than scaling model parameters for reasoning tasks (ICLR~2025).
\citet{liu2025art} show that no single test-time scaling strategy universally dominates but that performance scales monotonically with compute budget.
Chain-of-Visual-Thought \cite{chen2025covt} and related methods \cite{chen2025mcot} extend chain-of-thought reasoning to continuous visual tokens.
Our reasoning effort sweep provides direct empirical evidence for these theoretical results: GPT-5.4 improves from 12\% to 85\% to 91\%, and GPT-5.5 from 15\% to 92\% to 96\%, as reasoning effort increases from none to low to medium, with diminishing returns at higher budgets.


\section{Benchmark Design}
\label{sec:benchmark}

\subsection{Task Formulation}

Given a pixel-art maze image, the model must return a JSON object containing: the grid size, whether the start and goal are found, whether a path exists (\texttt{reachable}), the shortest path length, and the exact path as a list of directional moves (\texttt{U}, \texttt{D}, \texttt{L}, \texttt{R}).
A maze is scored as \emph{solved} only when all three conditions hold: (1)~reachability is correctly identified, (2)~the shortest path length is correct, and (3)~the returned path exactly matches one of the accepted shortest-path annotations.
No partial credit is awarded.

\subsection{Procedural Maze Generation}

We build a procedural generator that produces mazes with controlled parameters.
Each maze is defined by a grid size ($r \times c$), wall density $d \in [0, 0.55]$ (fraction of candidate cells converted to walls), trap count $t$ (impassable hazard tiles visually distinct from walls), and optional border walls (a wall ring around the outer edge).
Start and goal positions are randomized on opposite edges with a minimum Manhattan distance of $\lfloor(r+c)/3\rfloor$.

The generation algorithm places walls incrementally, verifying after each placement that the maze remains reachable (unless the maze is designated unreachable).
Traps are placed similarly with reachability checks.
Ground-truth shortest paths are computed via breadth-first search with multi-parent tracking, enumerating all optimal paths (capped at 50).
All mazes render as $1024 \times 1024$ pixel PNG images using procedurally generated pixel-art sprites across four visual palettes (forest, desert, dungeon, meadow).

\subsection{Dataset Structure}

The benchmark contains 110 mazes organized into nine groups:

\begin{itemize}
\itemsep0em
\item \textbf{Group~A: Diagnostic} (8 mazes). Empty or near-empty grids with straight-line paths. If a model fails here, the bottleneck is visual parsing, not reasoning.
\item \textbf{Group~B: Grid Scale} (15). Constant wall density (25\%), grid sizes from $5 \times 5$ to $13 \times 13$. Isolates the effect of spatial scale.
\item \textbf{Group~C: Wall Density} (15). Constant $9 \times 9$ grid, density swept from 0\% to 45\%. Isolates obstacle complexity.
\item \textbf{Group~D: Trap Ablation} (12). Six matched pairs sharing the same random seed---one with traps, one without---isolating trap recognition.
\item \textbf{Group~E: Unreachable} (14). All unreachable, spanning $5 \times 5$ to $13 \times 13$. Tests false-positive rate for reachability claims.
\item \textbf{Group~F: Border Walls} (10). Five matched pairs with/without border walls. Tests whether visual framing affects parsing.
\item \textbf{Group~G: Combined Hard} (16). Large grids ($9 \times 9$--$13 \times 13$), high density, traps, and borders combined.
\item \textbf{Group~H: Palette Stress} (10). Same maze structure rendered in all four palettes. Tests visual style sensitivity.
\item \textbf{Group~X: Ultra-Hard} (10). $20 \times 20$ grids with 8--25 traps, 35--55\% wall density, and shortest paths of 28--42 steps.
\end{itemize}

Overall the dataset contains 79 reachable mazes (72\% of the 110) and 31 unreachable (28\%).
Shortest path lengths range from 4 to 42 moves (mean 13.5, median 12).


\section{Experimental Setup}
\label{sec:setup}

\paragraph{Models.}
We evaluate models from four providers:
\textbf{OpenAI}: GPT-5.4 and GPT-5.4-mini, each at three reasoning effort levels (none, low, medium), plus GPT-5.5 (released April~2026) at none, low, and medium effort;
\textbf{Anthropic}: Claude Opus~4.6, Sonnet~4.6, and Haiku~4.5, each at no-thinking and low-effort configurations, plus Claude Opus~4.7 (released April~2026) at low- and medium-effort;
\textbf{Google}: Gemini~3.1~Pro~Preview and Gemini~3~Flash~Preview;
\textbf{Alibaba/Qwen}: Qwen~3.5~Plus and Qwen~3.5~Flash via DashScope.
This yields 19 model configurations in total.

\paragraph{Protocol.}
Every configuration receives the same fixed prompt instructing JSON-only output with no tool use.
Images are sent as base64-encoded data URLs.
We disable structured output enforcement and tool calling across all APIs so that models must reason freely.
Failed JSON parses are retried up to twice.

\paragraph{Reasoning effort control.}
For OpenAI, we use the \texttt{reasoning\_effort} parameter (none/low/medium).
For Anthropic, we test both no-thinking (omitting the thinking configuration) and adaptive thinking with low effort.
We evaluate Claude~4.6-era models only at no-thinking and low-effort because the text-grid ablation (Section~\ref{sec:textgrid}) establishes that their failure is in visual extraction, not downstream search---additional reasoning tokens applied to a misidentified grid yield no improvement.
For Claude Opus~4.7 we additionally include a medium-effort run, since its substantially higher solve rate (29--31\% vs.\ 4\% for Opus~4.6) suggests the visual-extraction bottleneck has partially lifted and makes effort scaling on the image task a meaningful comparison.
Gemini and Qwen models use default configurations; notably, Gemini performs hidden internal reasoning (visible via \texttt{thoughtsTokenCount} in the API response) that cannot be disabled.

\paragraph{Reproducibility.}
The main experiments were conducted between March~20--23, 2026, with the Claude Opus~4.7 runs added on April~16, 2026 and the GPT-5.5 runs on April~24--25, 2026, shortly after each model's release. The following API model identifiers were used: \texttt{gpt-5.5}, \texttt{gpt-5.4}, and \texttt{gpt-5.4-mini} (OpenAI Responses API), \texttt{claude-opus-4-7}, \texttt{claude-opus-4-6}, \texttt{claude-sonnet-4-6}, and \texttt{claude-haiku-4-5-20251001} (Anthropic Messages API, version \texttt{2023-06-01}), \texttt{gemini-3.1-pro-preview} and \texttt{gemini-3-flash-preview} (Gemini REST API), and \texttt{qwen3.5-plus} and \texttt{qwen3.5-flash} (DashScope API).
Temperature was set to 0.0 for all non-thinking configurations; Anthropic requires temperature~1.0 when thinking is enabled.


\section{Results}
\label{sec:results}

\subsection{Main Results}

Table~\ref{tab:leaderboard} presents the full leaderboard ranked by solve rate on the 100-maze core set (excluding Group~X ultra-hard).

\begin{table}[t]
\centering
\small
\begin{tabular}{@{}lrrr@{}}
\toprule
\textbf{Model} & \textbf{Solved} & \textbf{Reach.\%} & \textbf{Lat.(s)} \\
\midrule
\textbf{GPT-5.5 (medium)} & \textbf{96/100} & \textbf{99} & 28.6 \\
GPT-5.5 (low) & 92/100 & 99 & 13.5 \\
GPT-5.4 (medium) & 91/100 & 95 & 38.1 \\
GPT-5.4 (low) & 85/100 & 92 & 23.8 \\
Gemini 3.1 Pro & 79/100 & 86 & 51.9 \\
Gemini 3 Flash & 53/100 & 82 & 32.3 \\
GPT-5.4-mini (medium) & 51/100 & 86 & 21.8 \\
GPT-5.4-mini (low) & 49/100 & 89 & 7.3 \\
Opus 4.7 (low) & 31/100 & 74 & 29.1 \\
Opus 4.7 (medium) & 29/100 & 72 & 39.7 \\
GPT-5.5 (none) & 15/100 & 74 & 2.4 \\
Qwen 3.5 Flash & 15/100 & 71 & 12.2 \\
GPT-5.4 (none) & 12/100 & 71 & 2.1 \\
Qwen 3.5 Plus & 11/100 & 78 & 23.7 \\
GPT-5.4-mini (none) & 8/100 & 79 & 1.3 \\
Sonnet 4.6 (none) & 6/100 & 70 & 16.9 \\
Opus 4.6 (low) & 4/100 & 67 & 21.3 \\
Opus 4.6 (none) & 4/100 & 71 & 19.8 \\
Haiku 4.5 (low) & 3/100 & 63 & 9.1 \\
Sonnet 4.6 (low) & 2/100 & 60 & 14.7 \\
Haiku 4.5 (none) & 2/100 & 68 & 9.4 \\
\bottomrule
\end{tabular}
\caption{Main leaderboard on the 100-maze core set. \textbf{Solved} requires correct reachability, correct shortest path length, and exact path match. \textbf{Reach.\%} is reachability accuracy. \textbf{Lat.} is average latency per maze.}
\label{tab:leaderboard}
\end{table}

The results reveal a clear hierarchy.
The newly released GPT-5.5 with medium reasoning takes the top spot at 96\%, followed by GPT-5.5~low (92\%), GPT-5.4~medium (91\%), GPT-5.4~low (85\%), and Gemini~3.1~Pro (79\%).
GPT-5.5 advances the OpenAI ceiling by 5 points over GPT-5.4 at matched effort while also raising reachability accuracy to 99\% (vs.\ 95\% for GPT-5.4~medium).
All Claude~4.6-era models---Opus~4.6, Sonnet~4.6, and Haiku~4.5---score between 2--6\% regardless of size or thinking configuration, and enabling low-effort thinking sometimes \emph{degrades} performance (Sonnet drops from 6\% to 2\%).
The subsequently released Claude Opus~4.7 changes this picture partially: it jumps to 31\% at low effort and 29\% at medium effort---a roughly $8\times$ improvement over Opus~4.6 at matched effort---but still trails GPT-5.5~low (92\%) and GPT-5.4~low (85\%) by a wide margin.
Curiously, Opus~4.7 also exhibits inverse effort scaling in this range: medium effort solves \emph{fewer} mazes than low while consuming 22\% more tokens, echoing the pattern seen in the 4.6-era Sonnet low/none comparison.

\subsection{Reasoning Effort Scaling}
\label{sec:scaling}

Table~\ref{tab:scaling} presents per-group solve counts across all 16 model configurations, revealing both the effect of reasoning effort and stark cross-provider differences.

\begin{table*}[t]
\centering
\small
\begin{tabular}{@{}ll rrrrrrrr r@{}}
\toprule
& & \multicolumn{8}{c}{\textbf{Solves per Group}} & \\
\cmidrule(lr){3-10}
\textbf{Model} & \textbf{Reason.} & \textbf{A} & \textbf{B} & \textbf{C} & \textbf{D} & \textbf{E} & \textbf{F} & \textbf{G} & \textbf{H} & \textbf{Total} \\
& & \tiny{/8} & \tiny{/15} & \tiny{/15} & \tiny{/12} & \tiny{/14} & \tiny{/10} & \tiny{/16} & \tiny{/10} & \tiny{/100} \\
\midrule
\textbf{GPT-5.5} & \textbf{med.} & 8 & 13 & 13 & 12 & 14 & 10 & 16 & 10 & \textbf{96} \\
GPT-5.5          & low     & 8 & 15 & 11 & 10 & 14 & 10 & 14 & 10 & \textbf{92} \\
GPT-5.5          & none    & 8 &  2 &  0 &  0 &  2 & 0 &  3 & 0 & 15 \\
GPT-5.4          & med.    & 7 & 14 & 13 & 12 & 13 & 9 & 13 & 10 & \textbf{91} \\
GPT-5.4          & low     & 8 & 13 & 10 & 11 & 12 & 8 & 13 & 10 & \textbf{85} \\
GPT-5.4          & none    & 8 &  1 &  1 &  1 &  1 & 0 &  0 &  0 & 12 \\
GPT-5.4-mini     & med.    & 7 & 11 &  8 &  6 &  7 & 3 &  5 &  4 & 51 \\
GPT-5.4-mini     & low     & 6 &  9 &  8 &  5 & 11 & 3 &  3 &  4 & 49 \\
GPT-5.4-mini     & none    & 5 &  0 &  0 &  1 &  2 & 0 &  0 &  0 &  8 \\
\midrule
Gemini 3.1 Pro   & default & 8 & 12 & 14 &  9 &  9 & 7 & 10 & 10 & \textbf{79} \\
Gemini 3 Flash   & default & 7 &  7 &  9 &  7 &  6 & 5 &  7 &  5 & \textbf{53} \\
\midrule
Qwen 3.5 Flash   & default & 8 &  1 &  2 &  1 &  2 & 0 &  1 &  0 & 15 \\
Qwen 3.5 Plus    & default & 6 &  1 &  4 &  0 &  0 & 0 &  0 &  0 & 11 \\
\midrule
Sonnet 4.6       & none    & 3 &  0 &  1 &  1 &  0 & 0 &  0 &  1 &  6 \\
Sonnet 4.6       & low     & 1 &  0 &  1 &  0 &  0 & 0 &  0 &  0 &  2 \\
Opus 4.6         & none    & 2 &  0 &  1 &  0 &  1 & 0 &  0 &  0 &  4 \\
Opus 4.6         & low     & 2 &  0 &  1 &  0 &  1 & 0 &  0 &  0 &  4 \\
Haiku 4.5        & none    & 1 &  1 &  0 &  0 &  0 & 0 &  0 &  0 &  2 \\
Haiku 4.5        & low     & 3 &  0 &  0 &  0 &  0 & 0 &  0 &  0 &  3 \\
\bottomrule
\end{tabular}
\caption{Per-group solve counts for all 19 model configurations. Groups: \textbf{A}=Diagnostic, \textbf{B}=Grid Scale, \textbf{C}=Wall Density, \textbf{D}=Trap Ablation, \textbf{E}=Unreachable, \textbf{F}=Border Walls, \textbf{G}=Combined Hard, \textbf{H}=Palette Stress. GPT-5.5~medium is the first configuration to clear Group~G (Combined Hard) at full 16/16. GPT-5.4's none$\to$low transition (+73 solves) is the largest single improvement; GPT-5.5 shows a similar +77 jump (15$\to$92). All Claude models remain at 2--6 regardless of reasoning effort. Gemini models achieve strong results via hidden internal thinking.}
\label{tab:scaling}
\end{table*}

For GPT-5.4, the transition from no reasoning to low effort produces a dramatic +73 solve improvement (12$\to$85), while increasing to medium adds only +6 (85$\to$91), exhibiting clear diminishing returns.
The hard group (G) plateaus at 13/16 at both low and medium, suggesting that the remaining failures require qualitatively different capabilities rather than more reasoning tokens.
GPT-5.5 shows the same scaling shape but at a higher ceiling: none$\to$low adds +77 (15$\to$92) and low$\to$medium adds another +4 (92$\to$96), with the hard group (G) finally clearing at 16/16 at medium effort---the first model in our benchmark to do so.

In contrast, all Claude models remain flat at 2--6 solves regardless of reasoning configuration---Sonnet actually \emph{degrades} from 6 to 2 with low effort enabled.
Gemini models, which perform hidden reasoning by default, achieve results between GPT-5.4's low and medium configurations without any user-controllable effort setting.
Qwen models show modest performance (11--15), with the smaller Flash variant slightly outperforming Plus---the only family where the smaller model does better.

\subsection{Diagnostic Analysis: Vision Works, Reasoning Doesn't}
\label{sec:diagnostic}

Group~A (diagnostic) serves as a critical control.
These eight mazes have zero or near-zero walls with straight-line paths---any model that can parse the grid should solve them.
Both GPT-5.4 and GPT-5.5 achieve 8/8 \emph{even without reasoning}, confirming that their vision encoders correctly identify the grid, start position, goal position, and tile types.
GPT-5.4's collapse from 8/8 on diagnostics to 1/15 on Group~B (which adds only 25\% wall density) demonstrates that the failure is entirely in path planning, not visual parsing; GPT-5.5 follows the same pattern (8/8 $\to$ 2/15 without reasoning).

Interestingly, Claude models struggle even on diagnostics: Opus solves only 2/8 and Haiku 1/8 without thinking.
This suggests that Claude's vision pipeline has additional limitations in grid parsing that compound with the reasoning deficit.

\subsection{Unreachable Detection}

Unreachable mazes probe a different failure mode: can models recognize when no path exists?
Without reasoning, all models exhibit a strong bias toward claiming reachability---GPT-5.4 produces 25/28 false positives (89\% false-positive rate).
With medium reasoning, this drops to 6/28 (21\%), and unreachable detection reaches 93\% recall.
This suggests that detecting impossibility is itself a reasoning-intensive task: the model must exhaustively verify that no path exists rather than optimistically reporting one.

\subsection{Thinking Token Efficiency}
\label{sec:efficiency}

Figure~1 and Table~\ref{tab:efficiency} report total tokens consumed (thinking + output) per solve, revealing order-of-magnitude differences in reasoning efficiency.

\begin{table}[t]
\centering
\small
\begin{tabular}{@{}lrrr@{}}
\toprule
\textbf{Model} & \textbf{Solved} & \textbf{Tot.~Tok.} & \textbf{Tok/Solve} \\
\midrule
\textbf{GPT-5.5 (medium)} & \textbf{96} & 209K & 2,172 \\
\textbf{GPT-5.5 (low)} & \textbf{92} & 101K & \textbf{1,096} \\
GPT-5.4 (medium) & 91 & 265K & 2,913 \\
GPT-5.4 (low) & 85 & 145K & 1,710 \\
Gemini 3.1 Pro & 79 & 731K & 9,250 \\
Gemini 3 Flash & 53 & 804K & 15,171 \\
GPT-5.4-mini (medium) & 51 & 503K & 9,872 \\
GPT-5.4-mini (low) & 49 & 152K & 3,105 \\
Opus 4.7 (low) & 31 & 409K & 13,189 \\
Opus 4.7 (medium) & 29 & 500K & 17,239 \\
GPT-5.5 (none) & 15 & 10K & 668 \\
Qwen 3.5 Flash & 15 & 238K & 15,835 \\
GPT-5.4 (none) & 12 & 9K & 782 \\
Qwen 3.5 Plus & 11 & 240K & 21,804 \\
GPT-5.4-mini (none) & 8 & 11K & 1,427 \\
Opus 4.6 (none) & 4 & 91K & 22,818 \\
\bottomrule
\end{tabular}
\caption{Thinking token efficiency. \textbf{Tot.~Tok.} is total thinking + output tokens across all 100 mazes. \textbf{Tok/Solve} is the average total tokens consumed per correctly solved maze. Lower is more efficient.}
\label{tab:efficiency}
\end{table}

GPT-5.5 at low effort is the new Pareto-optimal configuration: 92 solves at \textbf{1,096} tokens per solve---a 36\% reduction over the previous Pareto point (GPT-5.4~low, 1,710 tok/solve) while solving 7 more mazes.
Medium effort on GPT-5.5 buys another 4 solves (92$\to$96) at roughly twice the token cost (1,096$\to$2,172).
GPT-5.4~medium remains a strong configuration but is dominated by GPT-5.5 on both axes: GPT-5.5~low ties or beats GPT-5.4~medium's solve rate (92 vs.\ 91) at 38\% fewer tokens per solve.
Gemini~3~Flash consumes 7,186 thinking tokens per maze internally (visible via the API's \texttt{thoughtsTokenCount} field) but achieves only 53\% solve rate---14$\times$ more total tokens than GPT-5.5~low for 39 fewer solves.
Claude~4.6-era models are the least efficient, spending 22,000--30,000 tokens per solve on verbose but incorrect outputs; Opus~4.7 improves to 13,000--17,000 tokens per solve, narrowing but not closing the gap to GPT-5.5~low (1,096 tokens per solve).

A key revelation is that Gemini models perform hidden reasoning by default: Gemini~3~Flash uses a median of 7,861 thinking tokens per maze despite having no user-configurable reasoning toggle.
The Gemini--Claude gap, however, is not solely explained by thinking tokens.
As our qualitative analysis in Section~\ref{sec:analysis} shows, Claude models produce inaccurate grid extractions (wrong dimensions, misplaced walls), meaning they brute-force on a \emph{hallucinated} grid---a compounding failure where poor vision quality renders even extensive reasoning futile.

\subsection{Ultra-Hard Ceiling}
\label{sec:ultrahard}

Group~X ($20 \times 20$ grids, 8--25 traps, paths of 28--42 steps) tests the absolute ceiling of current models.
Table~\ref{tab:ultrahard} shows per-maze results for GPT-5.5 at medium effort---the best-performing configuration on the core set.

\begin{table}[t]
\centering
\small
\begin{tabular}{@{}lccrcr@{}}
\toprule
\textbf{Maze} & \textbf{GT} & \textbf{Pred.} & \textbf{Path} & \textbf{Solved} & \textbf{Lat.} \\
\midrule
101 & \checkmark & --        & 40 $\to$ -- & $\triangle$ & 137s \\
102 & \checkmark & \checkmark & 30 $\to$ 30 & & 75s \\
103 & \checkmark & \checkmark & 42 $\to$ 42 & & 103s \\
104 & $\times$ & $\times$ & -- & \checkmark & 101s \\
105 & \checkmark & --        & 32 $\to$ -- & $\triangle$ & 87s \\
106 & \checkmark & \checkmark & 37 $\to$ 37 & \checkmark & 88s \\
107 & $\times$ & $\times$ & -- & \checkmark & 61s \\
108 & \checkmark & \checkmark & 28 $\to$ 28 & & 100s \\
109 & \checkmark & $\times$ & 41 $\to$ -- & & 77s \\
110 & $\times$ & $\times$ & -- & \checkmark & 87s \\
\midrule
\multicolumn{4}{@{}l}{Total solved} & \textbf{4/10} & 91s \\
\bottomrule
\end{tabular}
\caption{GPT-5.5 (medium) on the 10 ultra-hard 20$\times$20 mazes. \textbf{GT} = ground-truth reachability; \textbf{Pred.} = model prediction; \textbf{Path} = GT length $\to$ predicted length. \checkmark = correct, $\times$ = wrong, $\triangle$ = correct reachability but incomplete output (hit token limit). The model solves one reachable maze (37 steps) plus all three unreachable cases; it falsely declares the 41-step reachable maze unreachable and runs out of tokens on the 40- and 32-step mazes.}
\label{tab:ultrahard}
\end{table}

GPT-5.5 drops from 96\% on the core set to \textbf{4/10} on ultra-hard---a one-solve improvement over GPT-5.4~medium (3/10) but the same qualitative ceiling.
Average latency is 91 seconds (versus 28.6s on the core set), and 2/10 mazes hit the maximum output token limit (8,192 tokens).
GPT-5.5 cleanly solves all three unreachable ultra-hard mazes (full JSON output, correct reach); GPT-5.4 medium also identified them as unreachable but hit the token limit on two of them ($\triangle$) and was credited with only one (107).
Beyond that, GPT-5.5 still falsely declares the 41-step reachable maze unreachable and produces wrong paths on the 28-, 30-, and 42-step reachable mazes despite getting their lengths right.
The pattern is consistent: longer reachable paths exceed the brute-force budget even for the strongest model.

This provides direct evidence that models solve mazes through serial enumeration bounded by token budget: when the path exceeds what can be brute-forced within the thinking allocation, the model fails---and the failure mode is the same for both GPT-5.4 and GPT-5.5.


\section{Analysis and Discussion}
\label{sec:analysis}

\paragraph{A universal two-stage strategy.}
Examination of model outputs reveals that \emph{all} models---regardless of provider or accuracy---follow the same two-stage strategy:
\textbf{(1)~Image-to-grid translation}: the model converts the visual maze into a textual row-column matrix, and
\textbf{(2)~Serial path enumeration}: the model attempts to trace paths step-by-step through this textual grid.
What differentiates high-performing models from low-performing ones is primarily the quality of Stage~1, not the sophistication of Stage~2.

Claude models expose this strategy clearly because their reasoning traces are visible in the output.
On \texttt{gen\_maze\_014} (an 8$\times$8 grid), Opus~4.6 outputs a full grid transcription before searching:

\vspace{2pt}
{\small\ttfamily
\noindent Row 0: W, open, W, W, open, open, ...G\\
Row 1: W, S, open, W, W, W, open, ...\\
\mbox{[...maps all 10 rows...]}\\
Path: R,R,R,...(1,3) wait, (1,3) is wall.\\
Alternative: (1,1)→(1,2)→(2,2)→(2,3)→...\\
That's: R, D, R, R, R, R, R, R, R, U, U = 11\\
Shorter: ... = 11 moves. Or: ... = 15, longer.\\
Stick with 11.}
\vspace{2pt}

This is textbook brute-force search in natural language: try a path, hit a wall, backtrack, try another, count steps, compare.
Notably, our prompt explicitly instructs models not to ``use any external tools, code, search, calculators, or graph-search programs''---yet the models' only available strategy is to \emph{simulate} a graph-search algorithm (BFS) in natural language tokens, step by step.
Critically, both Opus and Sonnet \textbf{misidentify the grid as 10$\times$10} (it is actually 8$\times$8), leading them to reason over a hallucinated grid and produce incorrect paths.
GPT-5.4 correctly identifies 8$\times$8 and Gemini~3~Flash also reports the correct grid size---their Stage~1 is more accurate, which makes their Stage~2 brute-force search succeed on the correct grid.

\paragraph{The performance gap is in vision, not reasoning strategy.}
This two-stage analysis reframes the cross-provider performance differences.
Claude models do not fail because they use a worse \emph{reasoning} strategy---they use the same enumerate-and-backtrack approach as GPT-5.4.
They fail because their \emph{grid extraction} is unreliable: wrong grid dimensions, mislocated walls, and hallucinated openings.
On the trivially empty diagnostic maze \texttt{gen\_maze\_001} (5$\times$5, zero walls), Sonnet~4.6 reports a ``6$\times$6 grid'' instead of the correct 5$\times$5 and computes a path of length~4---the correct length, but from a misidentified grid.
Opus~4.6 reports ``7$\times$7'' and returns a path of length~5, both incorrect (the grid is 5$\times$5 with shortest path~4).

On the unreachable maze \texttt{gen\_maze\_057}, this vision deficit is catastrophic: Opus maps a 9$\times$9 grid as 11$\times$11 with incorrectly placed walls, then brute-forces a 14-step path \emph{through cells that are actually walls}---confidently declaring the maze reachable when it is not.
GPT-5.4 without reasoning correctly reports this maze as unreachable in 2.6 seconds, demonstrating superior visual parsing even without any chain-of-thought.

\paragraph{High accuracy still means brute-force.}
The crucial insight is that even the best-performing models---GPT-5.5 at 96\%, GPT-5.4 at 91\%, Gemini~3.1~Pro at 79\%---are still brute-forcing.
They simply brute-force on a \emph{correct} grid.
The evidence is threefold:
(1)~Performance scales with thinking token budget, not with any spatial heuristic---adding more tokens yields more solves, in a pattern consistent with longer serial search; this holds for GPT-5.5 (15$\to$92$\to$96) just as it does for GPT-5.4.
(2)~On ultra-hard 20$\times$20 mazes, GPT-5.5~medium still bottoms out at 4/10, with the same failure mode as GPT-5.4---hitting the token limit on the longest reachable paths and defaulting to ``unreachable.''
A human traces the same maze visually in seconds.
(3)~Token efficiency varies by 14$\times$ across providers (1,096 for GPT-5.5~low vs.\ 15,171 for Gemini~3~Flash), consistent with different search strategies expressed in text but inconsistent with human-like spatial planning, which should not require proportionally more computation for the same visual structure.

This is fundamentally different from human maze solving.
Humans engage spatial perception directly---tracing paths with their eyes, recognizing dead ends holistically, and planning routes without step-by-step enumeration.
Models convert vision to language, then search through language.
The conversion step is lossy (especially for Claude), and the search step is computationally expensive relative to the task's intrinsic difficulty.

\paragraph{Can prompting induce visual reasoning?}
A natural question is whether the brute-force strategy is a prompt artifact---perhaps the standard prompt implicitly encourages grid-based reasoning.
We test this with a modified ``visual-intuition'' prompt that explicitly instructs: \emph{``Do NOT convert the maze into a text grid, matrix, or row/column representation. Do NOT perform step-by-step BFS, DFS, or any graph-search algorithm in text. Instead, solve this the way a human would: look at the image, visually trace the walkable path.''}

On 20 representative mazes, GPT-5.4 (low) scores 17/20 with the visual prompt versus 19/20 with the standard prompt---slightly \emph{worse}, not better.
More revealing is what happens when we apply the visual prompt to Claude Sonnet~4.6, whose reasoning traces are visible.
Despite the explicit prohibition, Sonnet immediately falls back to grid enumeration:

\vspace{2pt}
{\small\ttfamily
\noindent Let me map out the walls (brown brick tiles)\\
and open paths (tan/sand tiles):\\
Row 0 (top): walls at col 0, col 2, then open\\
from col 3-8, wall at col 8-9 area\\
Row 1: Start at col 1, wall at col 1-2 area,\\
wall at col 3-5 area, open right side\\
\mbox{[...continues mapping all rows...]}}
\vspace{2pt}

In this setting, the model does not comply with the instruction to reason visually.
Instead, it falls back to converting the image into a textual grid representation and then enumerating paths through it---exactly the brute-force strategy the prompt forbids.
Crucially, the model never acknowledges this limitation---it silently violates the constraint rather than reporting that it cannot solve the task under the given restrictions.
This raises a broader concern about instruction compliance: when a model lacks the capability to follow a constraint, it proceeds with a forbidden strategy rather than failing gracefully.

\paragraph{Text-grid ablation confirms vision is the bottleneck---for both Claude generations.}
To directly test whether Claude's failure is in vision or reasoning, we bypass the image encoder entirely: we provide the model with the \emph{correct text grid} (using \texttt{S}, \texttt{G}, \texttt{.}, \texttt{\#}, \texttt{T} symbols) instead of the maze image.
We run this ablation on Sonnet~4.6 and on the newly released Opus~4.7 to test whether the partial image-input improvement in the 4.7 generation reflects better vision, better reasoning, or both.
Table~\ref{tab:textgrid} shows the results.

\begin{table}[t]
\centering
\small
\begin{tabular}{@{}llrrr@{}}
\toprule
\textbf{Model} & \textbf{Input} & \textbf{Solved} & \textbf{Tok/Solve} & \textbf{Lat.} \\
\midrule
Sonnet 4.6 & Image (none)    &  6/100 & 14,313 & 16.9s \\
Sonnet 4.6 & Text grid (low) & 80/100 &  3,222 & 34.8s \\
\midrule
Opus 4.7   & Image (low)     & 31/100 & 13,189 & 29.1s \\
Opus 4.7   & Text grid (low) & \textbf{90/100} & \textbf{2,285} & 18.7s \\
\bottomrule
\end{tabular}
\caption{Image vs.\ text grid on 100 mazes for two Claude generations. Bypassing vision yields $13\times$ more solves for Sonnet~4.6 and $2.9\times$ more for Opus~4.7. Opus~4.7 on text grids reaches 90\%---matching GPT-5.4~medium on images (91\%) at \emph{lower} tokens per solve (2{,}285 vs.\ 2{,}913).}
\label{tab:textgrid}
\end{table}

With low-effort reasoning on the text grid, Sonnet~4.6 solves \textbf{80/100} mazes---a 13$\times$ improvement over image input (6/100).
The same ablation on Opus~4.7 pushes performance to \textbf{90/100}, matching GPT-5.4~medium on images (91/100) and surpassing GPT-5.4~low (85/100) at a better tokens-per-solve figure (2,285 vs.\ 2,913 for GPT-5.4~medium, and 1,710 for GPT-5.4~low).
Token efficiency also improves dramatically in both cases: Sonnet drops from 14,313 to 3,222 tokens per solve (4.4$\times$), and Opus~4.7 drops from 13,189 to 2,285 (5.8$\times$).

The Opus~4.7 result is especially informative.
On images, Opus~4.7 already outperforms the 4.6-era Claude models by roughly $8\times$ (31\% vs.\ 4\%), suggesting the 4.7 generation has meaningfully improved grid extraction.
Yet its text-grid score (90\%) exceeds its image score (31\%) by a further 59 points---showing that even the improved 4.7 vision pipeline remains the dominant bottleneck on this task.
Across \emph{both} Claude generations we test, the reasoning engine is competitive with the best models when given accurate spatial input; the gap on images lives almost entirely in the image-to-grid translation stage---the same brute-force search that fails on a hallucinated grid succeeds on the correct one.
\label{sec:textgrid}

\paragraph{Within a generation, size does not buy spatial reasoning---but the generation does.}
Within the Claude~4.6 generation, Opus~4.6 solves only 4/100---fewer than GPT-5.4 with no reasoning (12/100), and tied with Haiku~4.5.
This is consistent with Claude's documented gap on MMMU \cite{yue2024mmmu} (80.7\% vs.\ GPT-5's 85.4\% and Gemini~3~Pro's 81.0\%).
The 4.7 generation, however, substantially closes that gap: Opus~4.7 reaches 31\% on images and 90\% when given the text grid, indicating that the image-to-grid pipeline has improved markedly between Claude generations, even though the underlying reasoning strategy (brute-force enumeration on an inferred grid) remains the same.

\paragraph{Hidden reasoning explains part of the Gemini advantage.}
Gemini outperforms the Claude~4.6 generation (53--79\% vs.\ 2--6\%) in part because it performs hidden reasoning by default: Gemini~3~Flash uses a median of 7,861 thinking tokens per maze, while 4.6-era Claude reports zero.
Combined with more accurate grid extraction, this enables effective brute-force search on correct grids.
Opus~4.7 narrows this gap meaningfully (31\% on images, 90\% on text grids), but the cross-provider ordering on image inputs still tracks vision quality more than raw reasoning capacity.


\section{Conclusion}
\label{sec:conclusion}

We introduced \textsc{MazeBench}, a benchmark of 110 procedurally generated visual mazes that probes not just \emph{whether} multimodal models can solve spatial tasks, but \emph{how} they do so.

Our main message is simple: \textbf{high accuracy on visual planning tasks can be misleading}.
The strongest models in our benchmark do solve many mazes, but they do not appear to solve them the way humans do.
Instead, they first translate pixels into a textual grid and then search for a path in token space, step by step.
This strategy is computationally expensive, brittle under scale, and qualitatively different from rapid visual path tracing.

This distinction matters for evaluation.
GPT-5.5 reaches 96\% on the 100-maze core set and GPT-5.4 91\%, yet both still rely on serial token-level search and collapse on ultra-hard mazes when the search no longer fits within the token budget---GPT-5.5~medium solves only 4/10 ultra-hard mazes despite cutting tokens-per-solve by 36\% on the core set.
Claude models expose the same search strategy even more clearly, but often on incorrect grid extractions; when we replace the image with the correct text grid, Sonnet~4.6 jumps from 6\% to 80\% and Opus~4.7 from 31\% to 90\%, showing that weak performance can come from poor visual extraction while strong performance can still come from brute-force planning on a correct grid.

Taken together, our results suggest that benchmark scores should not be read as evidence of human-like spatial understanding.
They are evidence only that a model can eventually produce the correct answer under a particular compute budget.
To measure genuine multimodal progress, we need evaluations that track not just correctness, but also \emph{strategy}, \emph{efficiency}, and failure mode.

\section*{Limitations and Future Work}

The benchmark uses procedurally generated pixel-art mazes and proprietary API-based models, and cross-provider reasoning controls are not perfectly comparable; future work should extend the setup to open models, human timing baselines, and broader visual planning domains.

\bibliography{references}

\clearpage
\appendix
\onecolumn

\section{Complete Maze Dataset by Group}
\label{app:mazes}

\begin{figure}[ht]
\centering
\includegraphics[width=0.85\textwidth]{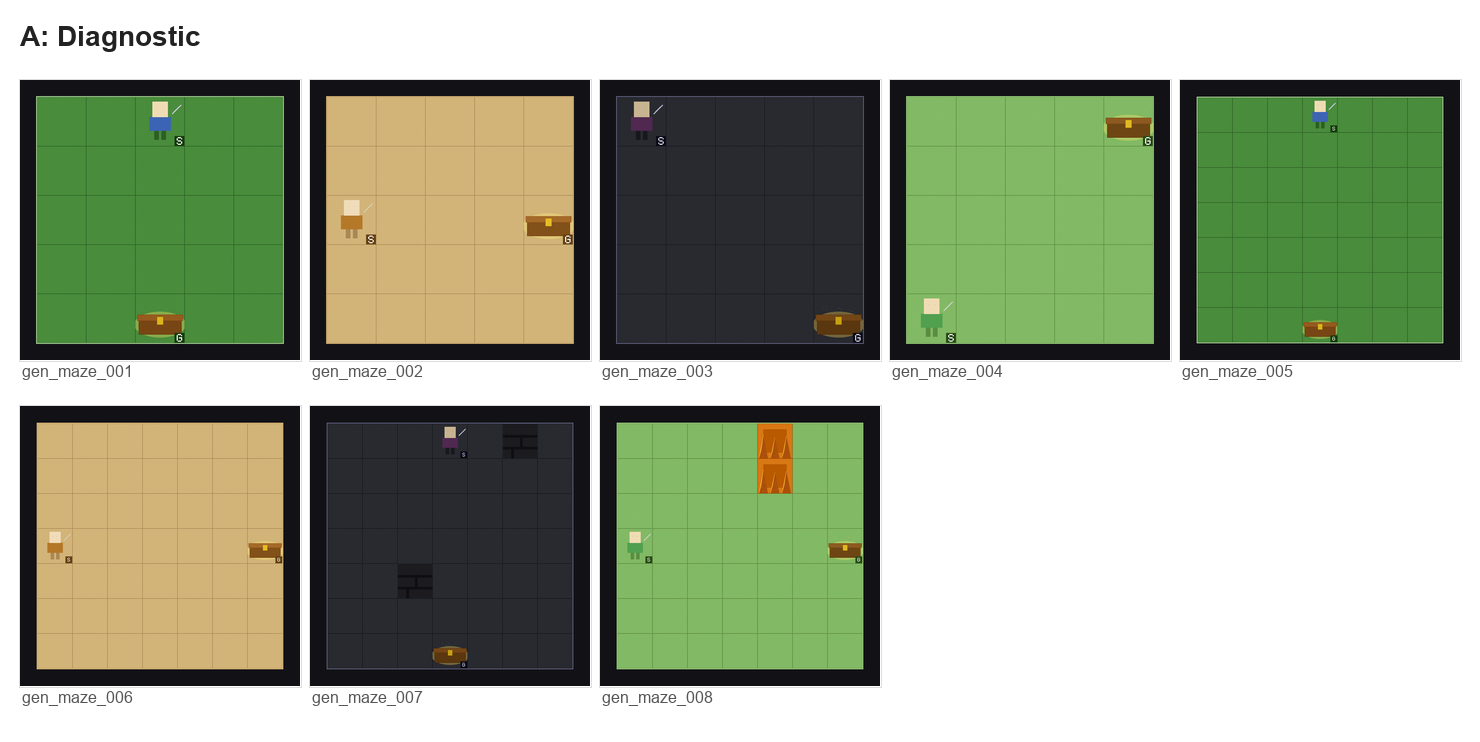}
\caption{Group~A: Diagnostic (8 mazes). Empty or near-empty grids with trivial straight-line paths.}
\label{fig:app_a}
\end{figure}

\begin{figure}[ht]
\centering
\includegraphics[width=0.85\textwidth]{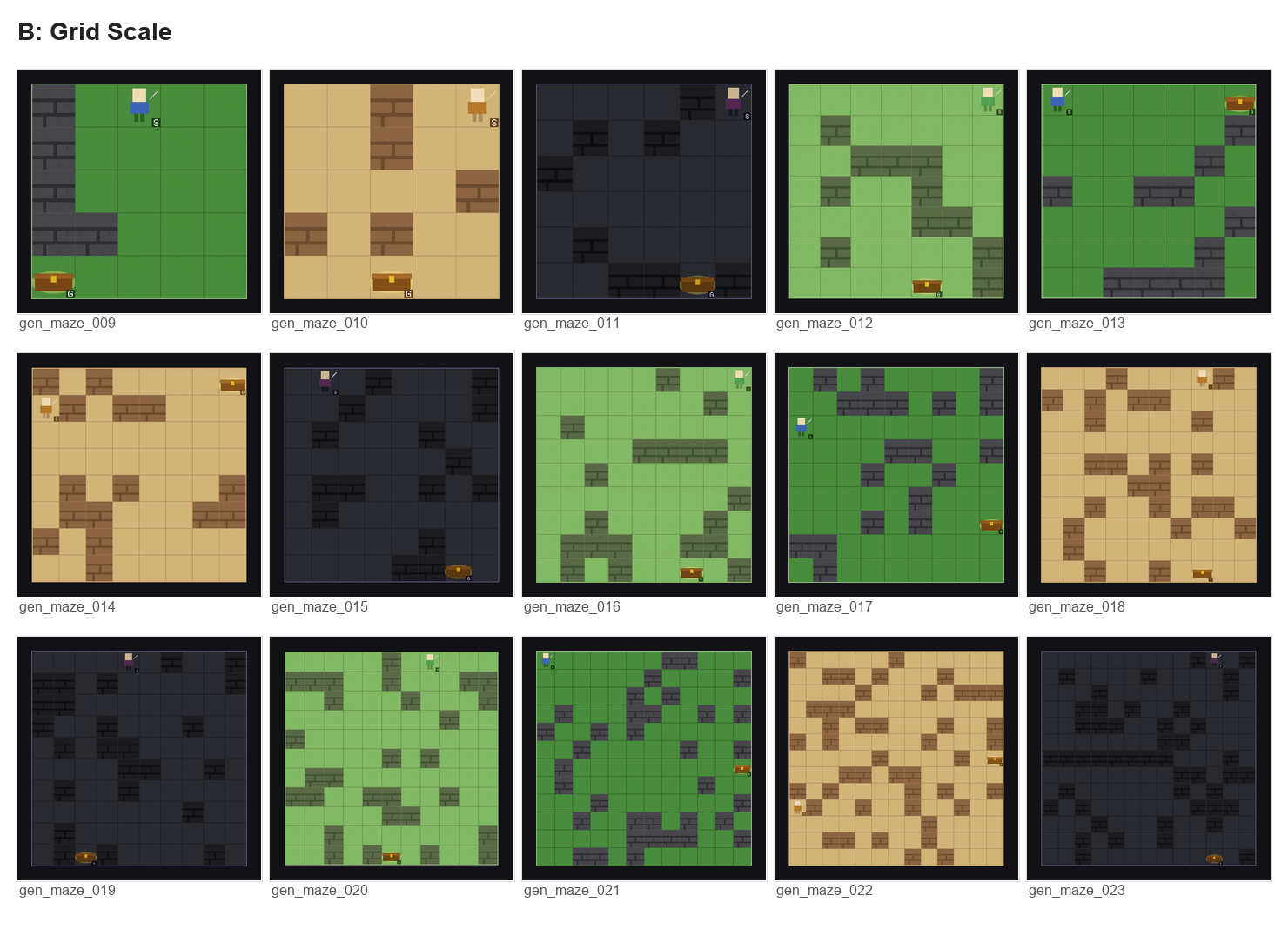}
\caption{Group~B: Grid Scale (15 mazes). Constant 25\% wall density, grid sizes from $5 \times 5$ to $13 \times 13$.}
\label{fig:app_b}
\end{figure}

\begin{figure}[ht]
\centering
\includegraphics[width=0.85\textwidth]{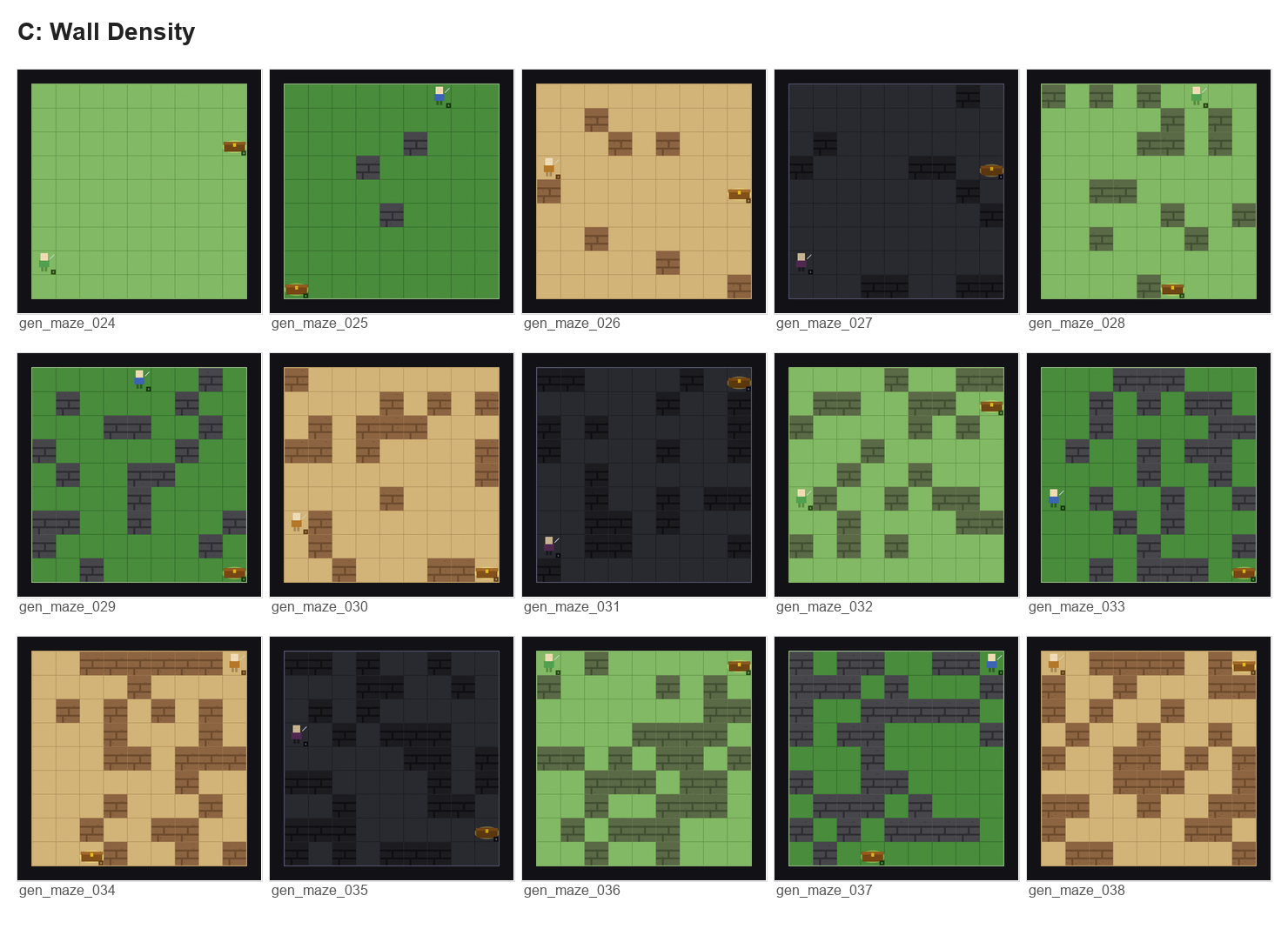}
\caption{Group~C: Wall Density (15 mazes). Constant $9 \times 9$ grid, density from 0\% to 45\%.}
\label{fig:app_c}
\end{figure}

\begin{figure}[ht]
\centering
\includegraphics[width=0.85\textwidth]{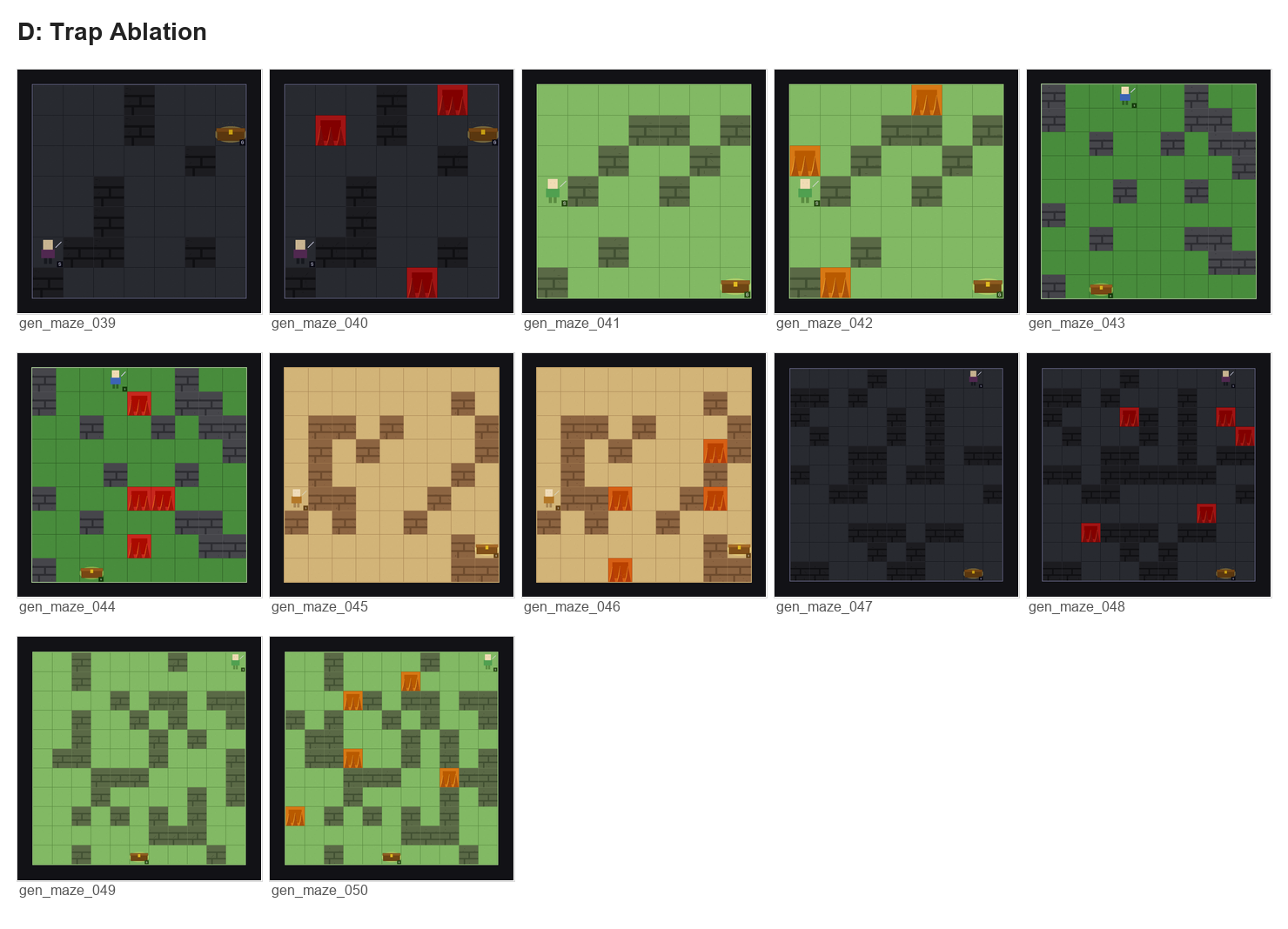}
\caption{Group~D: Trap Ablation (12 mazes). Six matched pairs---control (no traps) and treatment (with traps)---sharing the same random seed.}
\label{fig:app_d}
\end{figure}

\begin{figure}[ht]
\centering
\includegraphics[width=0.85\textwidth]{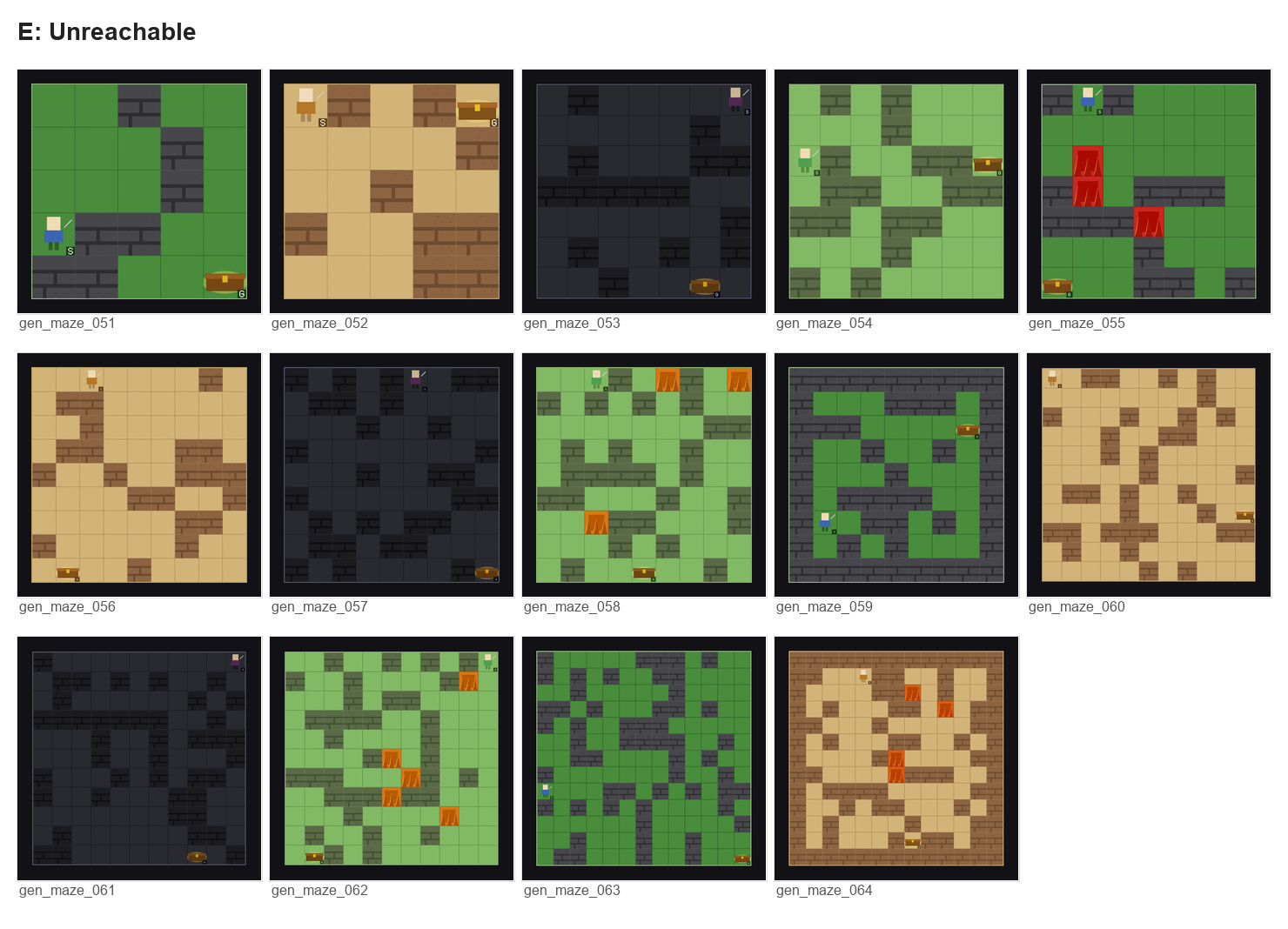}
\caption{Group~E: Unreachable (14 mazes). All mazes have no valid path from start to goal.}
\label{fig:app_e}
\end{figure}

\begin{figure}[ht]
\centering
\includegraphics[width=0.85\textwidth]{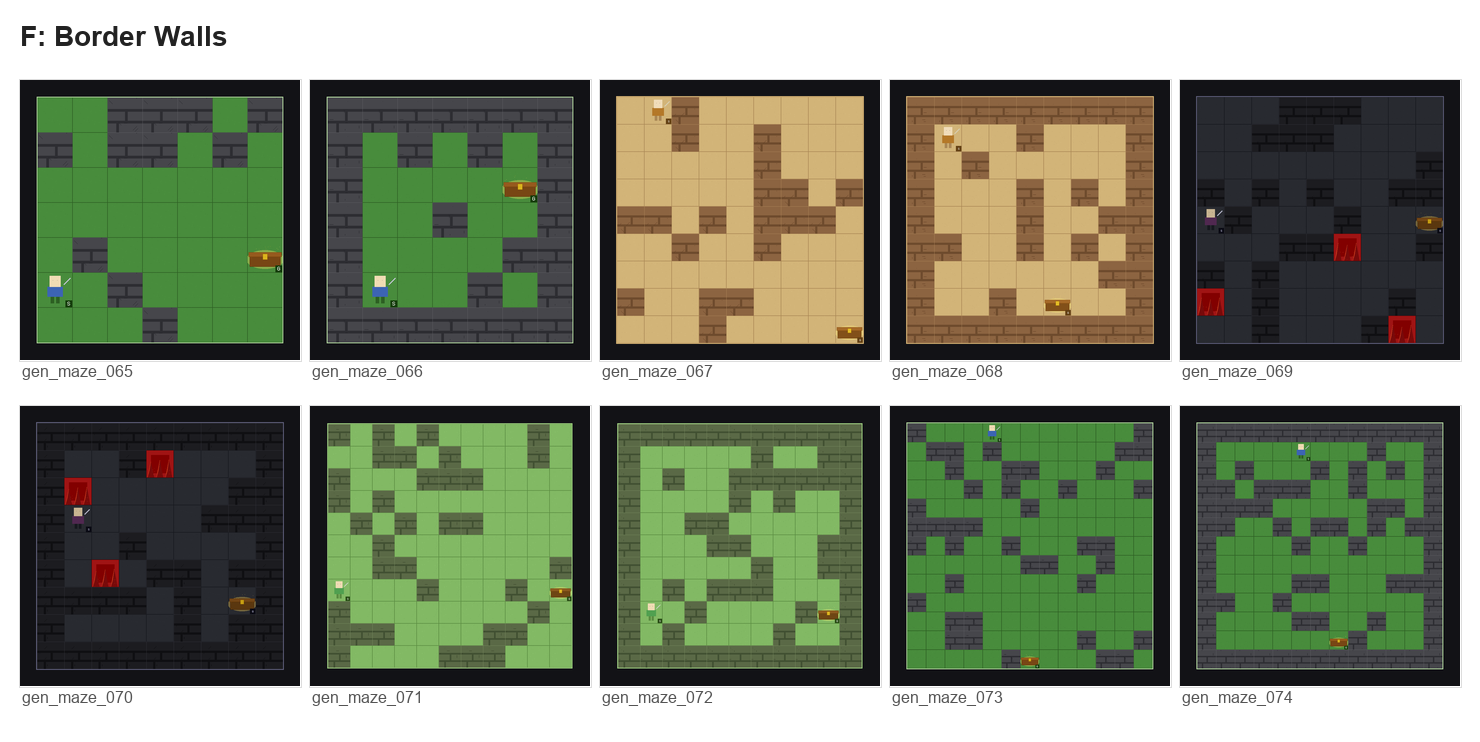}
\caption{Group~F: Border Walls (10 mazes). Five matched pairs with and without a wall border ring.}
\label{fig:app_f}
\end{figure}

\begin{figure}[ht]
\centering
\includegraphics[width=0.85\textwidth]{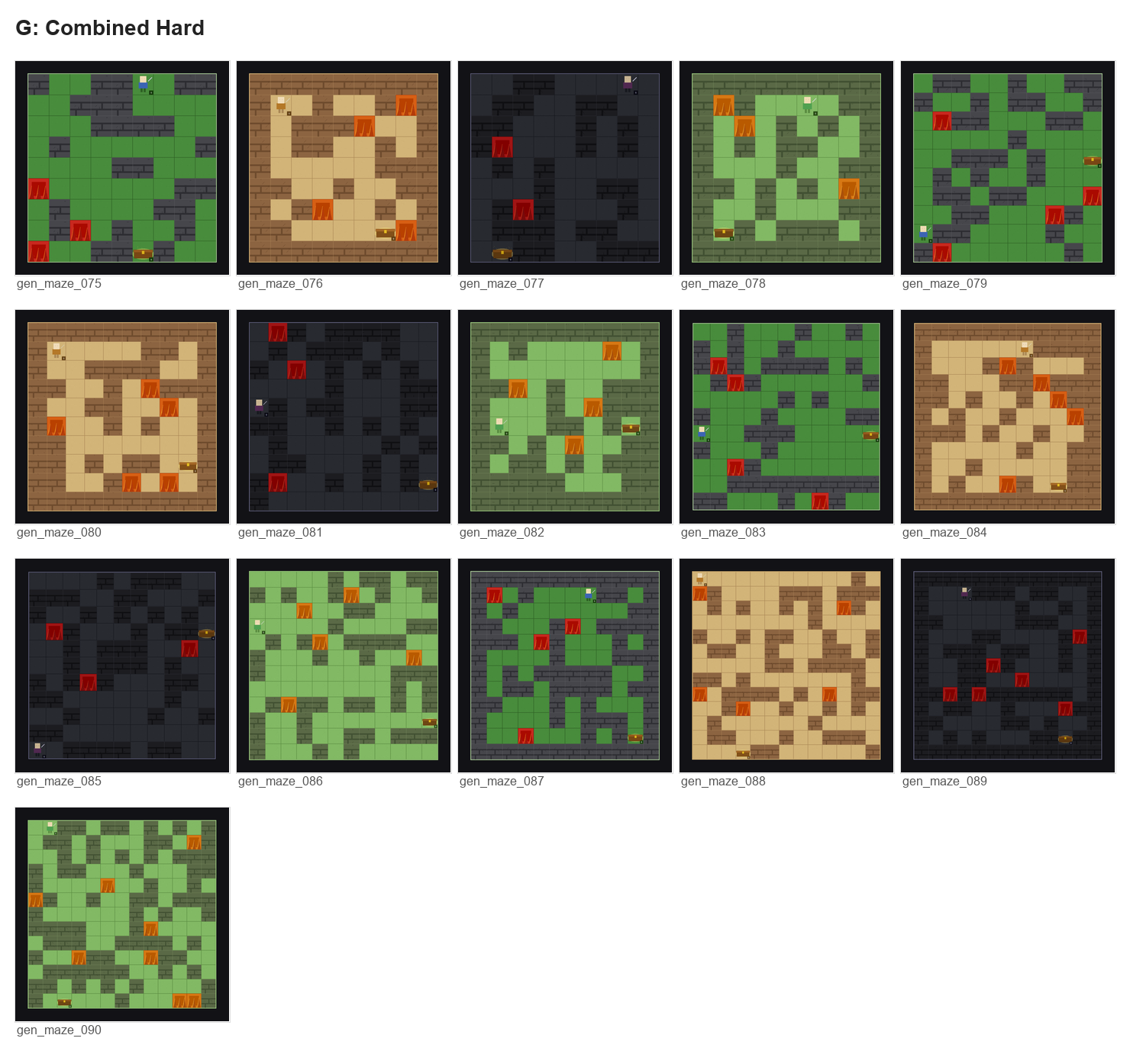}
\caption{Group~G: Combined Hard (16 mazes). Large grids with high wall density, traps, and borders.}
\label{fig:app_g}
\end{figure}

\begin{figure}[ht]
\centering
\includegraphics[width=0.85\textwidth]{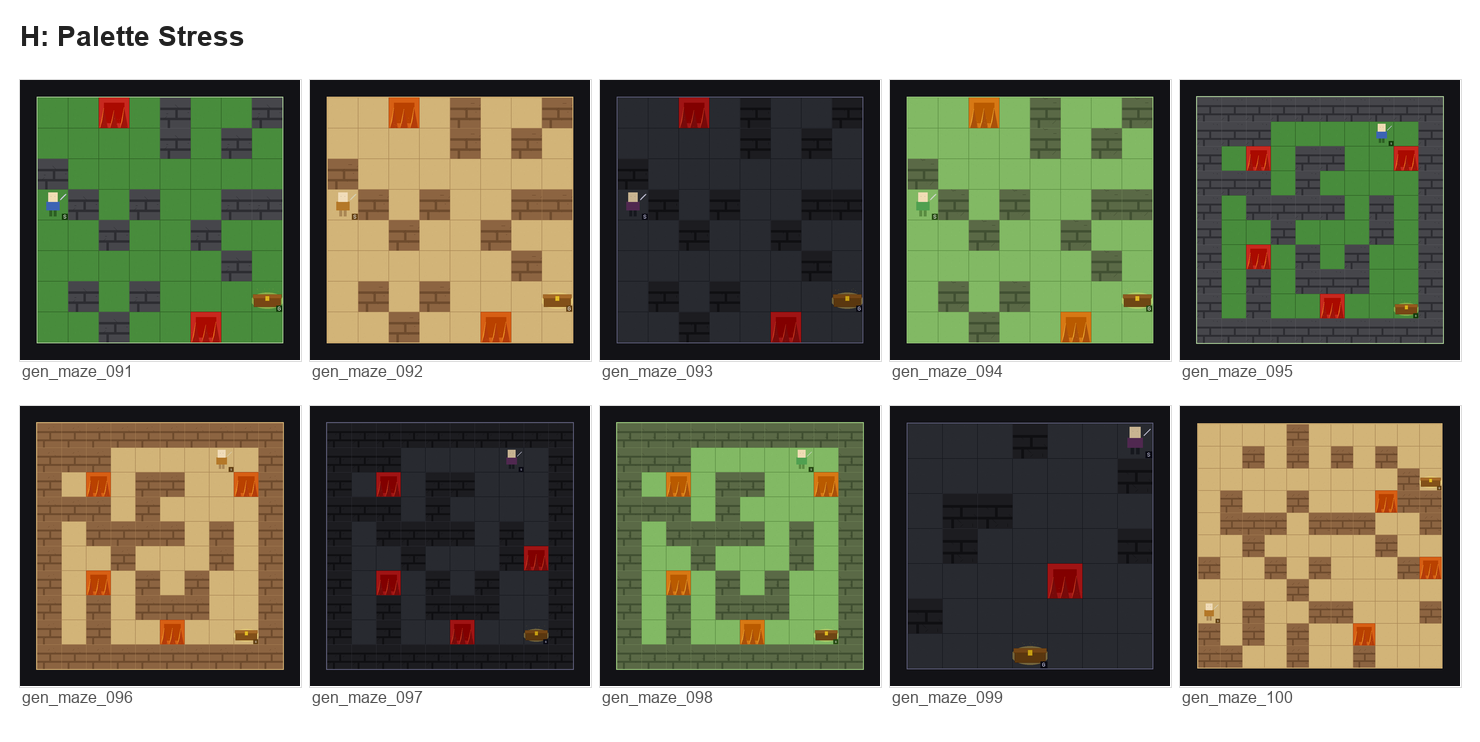}
\caption{Group~H: Palette Stress (10 mazes). Same maze structure rendered across four visual palettes.}
\label{fig:app_h}
\end{figure}

\begin{figure}[ht]
\centering
\includegraphics[width=0.85\textwidth]{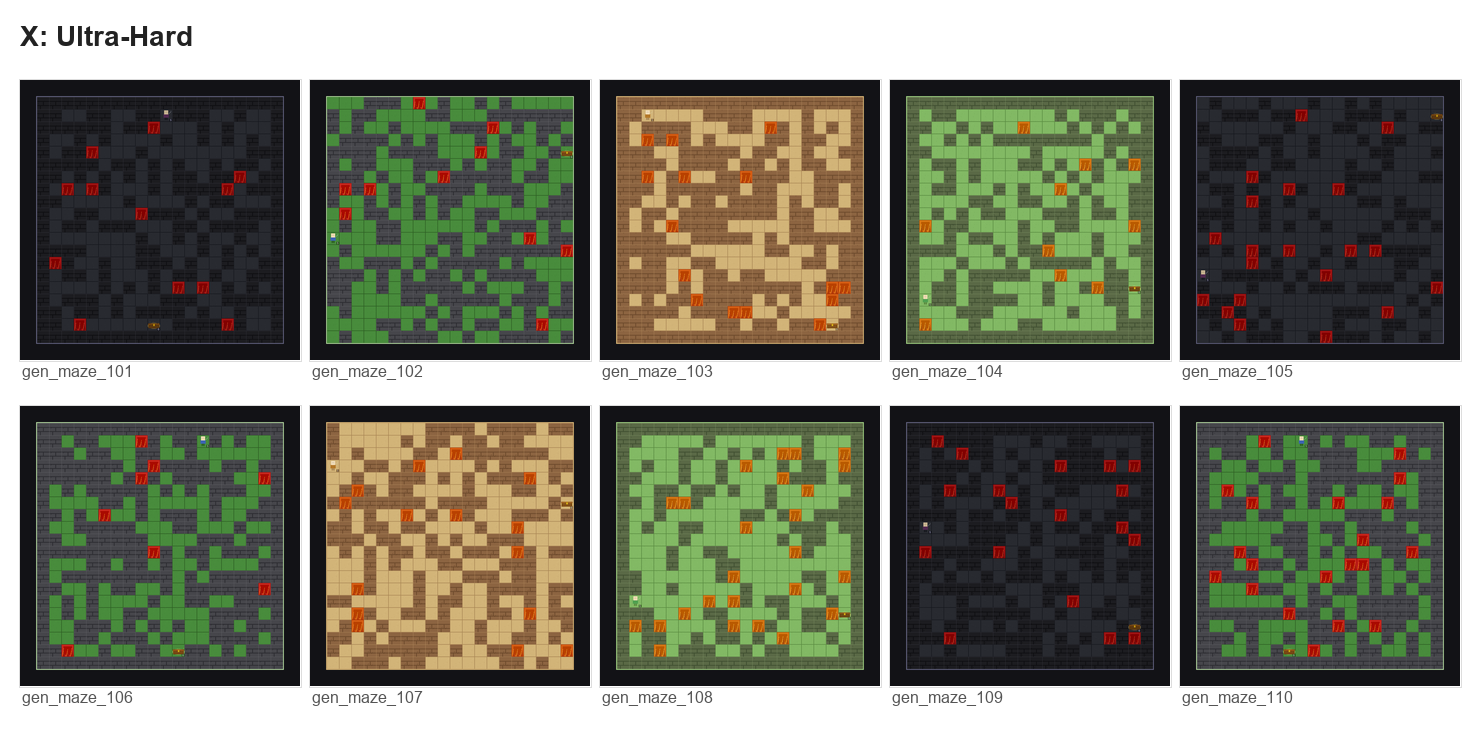}
\caption{Group~X: Ultra-Hard (10 mazes). $20 \times 20$ grids with 8--25 traps and shortest paths of 28--42 steps.}
\label{fig:app_x}
\end{figure}

Figures~\ref{fig:app_a}--\ref{fig:app_x} show every maze in the benchmark organized by experimental group.

\end{document}